\documentclass{article}
\usepackage{ijcai25}

\usepackage{times}
\usepackage{soul}
\usepackage{url}
\usepackage[hidelinks]{hyperref}
\usepackage[utf8]{inputenc}
\usepackage[small]{caption}
\usepackage{graphicx}
\usepackage{amsmath}
\usepackage{amsthm}
\usepackage{booktabs}
\usepackage{algorithm}
\usepackage{algorithmic}
\usepackage[switch]{lineno}
\usepackage{amssymb}
\usepackage{float}
\usepackage{tablefootnote}
\usepackage{appendix}

\usepackage{siunitx}

\DeclareSIUnit\fps{fps}

\usepackage{xcolor}

\newcommand{\tabref}[1]{Table~\ref{#1}}
\newcommand{\secref}[1]{Section~\ref{#1}}
\newcommand{\figref}[1]{Figure~\ref{#1}}
\newcommand{\apref}[1]{Appendix~\ref{#1}}

\title{Pay Attention to the Keys: Visual Piano Transcription Using Transformers}

\author{
Uros Zivanovic$^1$
\and
Ivan Pilkov$^2$\And
Carlos Cancino-Chacón$^{2}$\\
\affiliations
$^1$Universty of Trieste, Italy\\
$^2$Institute of Computational Perception, Johannes Kepler University Linz, Austria\\
\emails
uros.zivanovic@studenti.units.it,
ivan.pilkov@jku.at,
carlos\_eduardo.cancino\_chacon@jku.at
}

\begin{document}

\maketitle

\begin{abstract}

  Visual piano transcription (VPT) is the task of obtaining a symbolic representation of a piano performance from visual information alone (e.g., from a top-down video of the piano keyboard).
  In this work we propose a VPT system based on the vision transformer (ViT), which surpasses previous methods based on convolutional neural networks (CNNs).
  Our system is trained on the newly introduced \emph{R3} dataset, consisting of ca.~31 hours of synchronized video and MIDI recordings of piano performances.
  We additionally introduce an approach to predict note offsets, which has not been previously explored in this context.
  We show that our system outperforms the state-of-the-art on the PianoYT dataset for onset prediction and on the \emph{R3} dataset for both onsets and offsets.
  
\end{abstract}

\begin{figure*}[ht!]
    \centerline{
        \includegraphics[width=1.9\columnwidth]{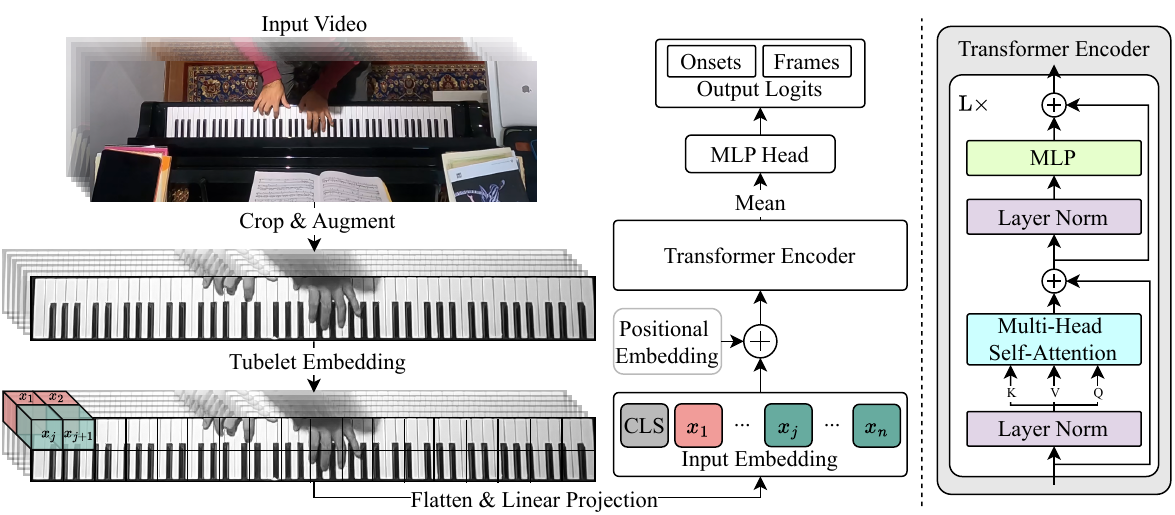}
    }
    \caption{Overview of the model architecture.}
    \label{fig:overall_architecture}
\end{figure*}

\section{Introduction}\label{sec:introduction}

Automatic music transcription (AMT) is the process of converting a music recording (typically an audio recording) into a symbolic representation (such as MIDI or sheet music) using computational methods~\cite{8588423}.
AMT is a cross-modal task that bridges audio and symbolic music processing and is a fundamental task in music information retrieval (MIR).
The generated symbolic music representation is often inspired by MIDI, with notes represented by their onset (start time), offset (release time), pitch, and velocity (loudness).
Various MIR tasks, such as music search (indexing, recommendation), automatic music accompaniment, and music generation, benefit from the symbolic music representation produced by an AMT algorithm~\cite{8588423,10.1561/1500000042}.
AMT systems also facilitate the creation of new datasets~\cite{zhang2022atepp,Kong-2022}, which can support a range of MIR tasks.

Although audio-based methods have been the primary focus of past AMT research, transcription from visual information, either alone or in combination with audio, presents an interesting approach to exploring cross-modal mappings, particularly in linking finely controlled gestures/movements with musical actions.
Learning these mappings is valuable not only for AMT itself but also as an intermediate representation for tasks that connect motion to music-making, such as music learning tools that could provide feedback during a performance.

In this work, we focus specifically on the case of automatic piano transcription (APT) for silent video.
A top-down video can provide a substantial amount of information about a piano performance, allowing for the direct visualization of note onsets, offsets, and pitch-values based on key-presses and releases. 
Videos with such setup are a popular way to showcase piano-related content on platforms such as YouTube.\footnote{See, e.g., \url{https://www.youtube.com/watch?v=xspj9uE5TQQ}}
Research into video-based APT has shown that visual input can supplement audio-based methods, helping models handle harmonics~\cite{https://doi.org/10.1049/cje.2015.07.027} and improving overall system performance in cases where the audio data is distorted~\cite{9053115}. 
Additionally, video can assist with tasks such as video--audio synchronization and it can provide a transcription solution in scenarios where no audio is available for a musical performance.

Previous research on visual piano transcription (VPT) has used convolutional neural networks (CNNs) for this task~\cite{9768275,9053115,https://doi.org/10.1049/cje.2015.07.027}. 
Recently, the video vision transformer (ViViT)~\cite{DBLP:conf/iccv/Arnab0H0LS21} has been shown to be competitive with state-of-the-art (SotA) CNNs in computer vision benchmarks.
This is leading to a shift in the computer vision space from CNNs to transformers.
Training a machine learning VPT system requires data in the form of videos and corresponding MIDI files. 
Current machine learning methods rely on datasets generated from audio-based APT for training and testing (i.e., using APT to transcribe the audio track in the video, and use the resulting MIDI file as the ground truth)~\cite{DBLP:conf/nips/SuLS20,9053115}. 
However, using these datasets may introduce artifacts due to limitations inherent in audio-based AMT.
In particular, recent work has highlighted the limitations of APT, primarily due to SotA APT methods being trained on the same dataset (the MAESTRO dataset~\cite{hawthorne2019maestro}). 
Additionally, audio-based APT is limited in accurately transcribing note durations for this purpose, due to pianists' use of the sustain pedal.~\cite{hu2024towards,marták2024quantifyingcorpusbiasproblem,10428040}. 
Furthermore, current VPT methods focus on predicting onsets and pitches only~\cite{9053115,suteparuk2014detection}, but not offsets (with the exception of \cite{DBLP:conf/nips/SuLS20}, which generates an intermediate piano-roll--like representation to generate audio).

The contributions of this paper are the following:
\begin{enumerate}
  \item We propose a system based on the vision transformer (ViT)~\cite{DBLP:conf/iclr/DosovitskiyB0WZ21} that is competitive with the SotA in VPT. The proposed system is illustrated in Figure \ref{fig:overall_architecture}.
  \item We investigate two approaches to predict offsets, namely predicting frames directly, as well as an approach based on the Onsets and Frames (OaF) APT model for audio \cite{DBLP:conf/ismir/HawthorneESRSRE18} and we compare the overall performance of the two for the purpose of our task.
  \item We introduce the \emph{R3} dataset, containing 895 synchronized and high quality 
video, audio, and MIDI recordings of two professionally trained pianists practicing various pieces, amounting to ca.~31 hours of music. 
  To the best of our knowledge, this is the largest dataset for the task, being almost twice the size as the previous largest dataset~\cite{9053115} (which consists of transcribed piano performances, not real ``performed'' MIDI files).
\end{enumerate}

Alongside our contributions, we also present the rationale behind key choices made during the design of our system.
These include the results of our experimentation in progressively adding data augmentation techniques when training the ViT, as well as our use of loss weights to deal with the sparsity of positive note events.
Lastly, we present the overall performance of our model and compare it directly to other SotA methods.

The structure of this work is as follows:
\secref{sec:related_work} reviews related work on visual transcription and multi-modal computer vision tasks.
\secref{sec:visual_transformers} describes our ViT-based transcription architecture.
\secref{sec:r3dataset} describes the R3 dataset.
\secref{sec:experimental_setup} describes the experimental setup.
\secref{sec:experiments} presents the results of our experiments.
\secref{sec:discussion} provides a discussion.
Finally, \secref{sec:conclusions} concludes the paper.

\section{Related Work}\label{sec:related_work}

In this section we discuss some related work on APT and transformer-based multi-modal computer vision tasks.

\subsection{Audio-based Automatic Piano Transcription}

Piano transcription is one of the most popular subtasks in AMT.
Its popularity is partially due to the relatively limited number of degrees of freedom for transcribing piano (i.e., needing onset and offset times, played note and potentially pedal), when compared with other instruments like violin or voice.
Furthermore, there are large datasets for audio-based piano transcription, including the MAESTRO dataset~\cite{hawthorne2019maestro}.
A comprehensive review of AMT is provided in~\cite{8588423}.

Early approaches for audio-based APT used a sliding window over the audio spectrogram, predicting whether each note was present in the current ``frame''~\cite{DBLP:conf/ismir/KelzDKBAW16}.
Such an approach is similar to the one taken by VPT models, which iterate over the frames in a video~\cite{9053115,suteparuk2014detection}.
The OaF model~\cite{DBLP:conf/ismir/HawthorneESRSRE18} was the first audio based AMT model to conduct dual-objective piano transcription, showing that it can be beneficial to model note onsets separately from frames.

CNNs have been the model of choice for spectrogram-based APT~\cite{DBLP:conf/ismir/HawthorneESRSRE18,DBLP:conf/ismir/KelzDKBAW16,Kong-2022}.
In recent years, however, transformer-based models been shown to work well for both multi-track transcription~\cite{51236} and single-track piano transcription~\cite{DBLP:conf/ismir/HawthorneSSME21}.
By using end-to-end machine learning architectures, these works move the problem from hand designing a system for AMT to creating a high-quality dataset and fitting a model to it.

\subsection{Video-based Automatic Piano Transcription}
Because of the easily discernible black and white keys of the piano, traditional computer vision algorithms are able to register a piano keyboard in an image and even identify separate keys.
This enables such tools to detect onsets on the piano with a high degree of accuracy~\cite{suteparuk2014detection}.
Such an approach is very difficult to apply to various lighting and environmental conditions, however.
It also struggles in cases where the piano keys may be slightly obscured, or when the piano is at an angle in the image.
Additionally, traditional computer vision tools are unable to capture and reason about complex movements across multiple frames in a video.

More recently, CNNs have been shown to work well for VPT.
By using a modified ResNet~\cite{7780459} model, \cite{9053115,DBLP:conf/nips/SuLS20} were able to make onset and pitch predictions directly on a cropped image of the piano.
\cite{9053115} showed that their vision-based model can compensate for poor quality audio, helping an audio-based AMT algorithm perform better.
There are also systems that are explicitly designed around using both video and audio simultaneously.
Such an audio-visual model is presented by \cite{9768275}.
In their approach, they use the video model primarily for eliminating harmonic errors.
This is done by tracking the pianists hands and eliminating any predicted notes that would be impossible to play from that position.
\cite{DBLP:conf/nips/SuLS20} present Audeo, an approach that generates audio by first generating an intermediate pianoroll-like representation, followed by a synthesis step.
The main component of this intermediate step, referred to as Video2Roll Net, is a ResNet model similar to the one used by \cite{9053115}, but enhanced with feature attention for improving visual cues at multiple scales.

\subsection{ViT for Video and Multi-modal Applications}

Since ViT showed that transformer architectures can outperform comparative CNNs in the ImageNet benchmark~\cite{DBLP:journals/ijcv/RussakovskyDSKS15}, there has been an increased interest in transformers for computer vision.
When adapting ViT to video, there have been numerous architectures proposed, although most focus on classification. 
SwinBERT~\cite{DBLP:conf/cvpr/LinLL0G0LW22} is a multi-modal architecture which is able to take a video clip and generate a text caption describing it.
While SwinBERT uses both a Swin transformer~\cite{DBLP:conf/iccv/LiuL00W0LG21} and a multimodal transformer encoder to generate the caption text, we are able to avoid needing a second model thanks to the simplicity of piano transcription compared to language.

Our proposed model is instead more similar to the spatio-temporal attention variant of ViViT~\cite{DBLP:conf/iccv/Arnab0H0LS21,DBLP:conf/icml/BertasiusWT21}.
This type of ViViT is a direct generalization of ViT to video, and has been utilized by the Video Masked Autoencoders~\cite{DBLP:conf/nips/TongS0022} for a variety of video classification tasks including fine-action detection.
The Video-Audio-Text transformer~\cite{DBLP:conf/nips/AkbariYQCCCG21} also utilizes the same method, but appends additional audio and text tokens to the input sequence to conduct multimodal self-supervised learning, generating representations that are useful for both video and audio classification.

\section{Visual Transcription Architecture}\label{sec:visual_transformers}

In this section we describe the details of our ViT-based transcription model.
For a visual overview of the system, see \figref{fig:overall_architecture}.

\subsection{Predicting Onsets and Offsets}

When considering methods for onset/offset prediction, the simplest choice is to conduct 88 classifications for each frame in the video, predicting what notes are present in each given frame.
These framewise predictions can then be concatenated into an array to form a prediction for the entire video.
A VPT method that utilizes this approach is the Audeo model~\cite{DBLP:conf/nips/SuLS20}, which additionally places an extra generative adversarial network on top of the video predictions.

\begin{figure}[t]
  \begin{center}
  \includegraphics[width=\linewidth]{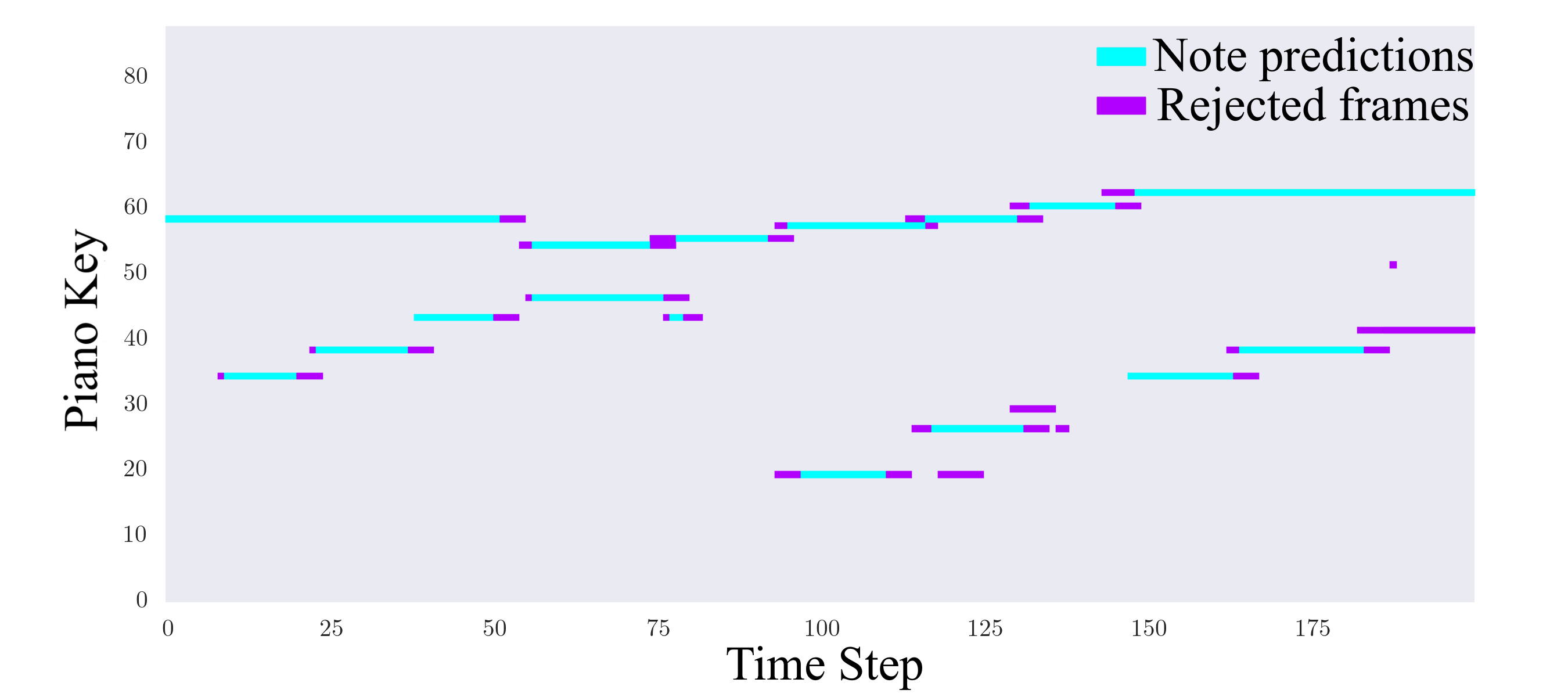}
  \end{center}
  \caption{An illustration of accepted and rejected frame predictions.}
  \label{fig:onsets_frames}
\end{figure}

Splitting a note up into distinct onset and frame predictions, enables us to reject spurious positive frames when they are missing a corresponding onset.
An example of this is visualized in \figref{fig:onsets_frames}.
Modeling onsets separately from the rest of the note is also closer to what is happening physically in the video, as pressing down a note and holding down a note are two different actions.

\paragraph{Inference.}
When performing inference over an entire video, we iterate over all frames and then concatenate the predictions to produce an array of shape
$$(88\times 2,\verb|number_of_frames|-\lceil\verb|window_size|/2\rceil\times 2).$$
Some frame predictions are missing at the start and end because we need enough frames to create a window.
An alternative to skipping the first and last few frames could be to pad the start and end of the video.
As there are 88 notes, to predict both onsets and frames we require 176 total predictions per frame.
Because we train our model on a temporal resolution of \SI{30}{\fps}, all predicted notes get assigned to their nearest frame, leading to a potential error of \SI{15}{\milli\second} even given perfect predictions. 
Achieving a high temporal accuracy is therefore a key area where VPT is more challenging than audio based AMT.

\subsection{Post-Processing}\label{sec:visual_transformers:post_processing}

After iterating through the video, the resulting array still contains some noise.
To remove this we apply a Gaussian filter along the time axis (similarly to \cite{9053115}) individually to the onset and frame predictions. For onsets we use $\sigma=0.2$ and a kernel radius of 8, while for frames we use $\sigma=0.8$ and a kernel radius of 4.
We then threshold the smoothed array using a value of 0.5.

To identify onsets, we find regions in the onset predictions which are positive for 1 or more consecutive time steps, and assign the corresponding onset to the time step with the highest predicted value in that region.
After finding an onset for a note, we then iterate over all positive frames that succeed that onset until we reach the end of the predicted frames. 
The final 4 frames are ignored in order to compensate for the Gaussian smoothing (which can cause false positives up to the size of the kernel radius).
At this point we have a completed piano roll, which can be converted directly into MIDI.
Although our model outputs both onsets and frames, we can also choose to utilize either one of them to predict only onsets or only frames.
We take this approach in \secref{sec:experiments:predictive} to evaluate how well frame predictions perform for predicting onsets and offsets.

\subsection{Vision Transformer Backbone}

We introduce a new backbone architecture for VPT built upon the ViT, which accepts a video clip sampled at \SI{30}{\fps} consisting of 6 frames with a resolution of $(64, 784)$, and outputs a vector of size $(176)$ containing onset and frame predictions for the third frame in the input window.
The architecture is visualized in \figref{fig:overall_architecture}.

In order to convert the input video clip to embeddings, we utilize the tubelet embedding method used in ViViT~\cite{DBLP:conf/iccv/Arnab0H0LS21}. 
Tubelet embeddings are a direct generalization to 3D of the ``patchification'' procedure used in ViT.
Defining a tubelet size of ($t\times h\times w$), we reshape a video with size ($T,H,W$) to: 

$$
\left(\biggl\lfloor\frac{T}{t}\biggr\rfloor\times \biggl\lfloor\frac{H}{h}\biggr\rfloor\times \biggl\lfloor\frac{W}{w}\biggr\rfloor,t\times h\times w\right).
$$

\noindent A linear layer is then applied to project the flattened tubelets $x_{i}\in \mathbb{R}^{t\times h\times w}$ to $z_{i}\in \mathbb{R}^{d}$, where $d$ is the embedding dimension.
After appending a learned \texttt{[CLS]} token to the start of the sequence and adding the positional embeddings, the tokens are passed through a standard transformer encoder.
While VPT approaches utilizing ResNets benefit from the addition of a ``slope vector'' to help the model identify the location of keys in the image~\cite{9053115,DBLP:conf/nips/SuLS20}, because ViT uses absolute positional embeddings, we do not require any such modifications.

As vision transformers have been shown to require extensive pre-training to achieve satisfactory results, we choose to utilize the high-quality distilled ViT-small weights provided by ~\cite{DBLP:conf/cvpr/WangHZTHWWQ23}.
The ViT-small model consists of 12 Transformer Encoder layers, an embedding dimension $d$ of 384, 6 attention heads, and an MLP hidden dimension of 1536.
The model was pre-trained on 16 frame long video clips with a resolution of (224$\times$224) and a tubelet size of (2$\times$16$\times$16).
We change the resolution of the model to (64, 784), and number of frames to 6.
Although this is a large change, we choose the resolution such that the number of tokens per frame stays the same: 

$$
\frac{H}{h}\times\frac{W}{w}=\frac{64}{16}\times\frac{784}{16}=\frac{224}{16}\times\frac{224}{16}=196.
$$

\noindent When adapting the positional embeddings to the new number of frames, we simply remove the section that used to correspond to the extra frames and keep the rest.
To generate onset and frame predictions, we follow the same procedure used during pre-training, taking the mean of the output embeddings and placing a linear head on top.
Therefore, we are able to utilize the pre-trained model with a different resolution and number of frames without requiring the re-initialization of any weights or rescaling of the positional embedding.

\section{Datasets}\label{sec:r3dataset}

\tabcolsep=0.13cm
\begin{table}[t]
  \begin{center}
    \begin{tabular}{ccccc}
      \toprule
                    & \multicolumn{2}{c}{No. Notes} & \multicolumn{2}{c}{Total Duration} \\
      Dataset       & Test   & Train   & Test  & Train  \\
      \midrule                  
      R3s           & 98,045  & 292,814 & 2.9h & 8.5h  \\
      R3x           & 119,725 & 479,629 & 3.6h & 15.8h \\
      R3 (Total)    & 217,770 & 772,443 & 6.6h & 24.3h \\
      \midrule
      PianoYT       & 33,535  & 369,922 & 1.7h & 17.2h \\
      \bottomrule
    \end{tabular}
  \end{center}
  \caption{Statistics on the number of notes and total video duration in the R3 (R3s + R3x) and PianoYT datasets.}
  \label{tab:dataset_statistics}
\end{table}

\begin{figure}[t]
  \begin{center}
  \includegraphics[width=\linewidth]{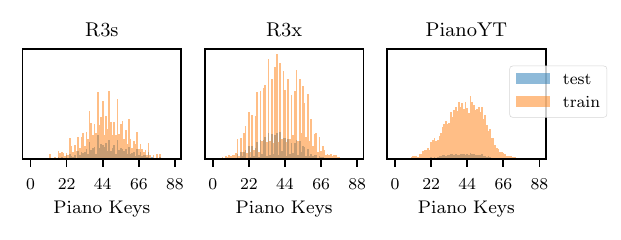}
  \end{center}
  \caption{Note distribution for the R3 and PianoYT datasets.}
  \label{fig:dataset_distribution}
\end{figure}

\subsection{R3}
In order to train our ViT-based model, we introduce a dataset consisting of ca.~31 hours (shown in \tabref{tab:dataset_statistics}) of recordings from 62 different piano practice sessions by two professionally trained pianists (each with an average of 14 years of formal music education).
The recordings come from work on the \emph{Rach3} dataset~\cite{DBLP:conf/mmm/ChaconP24} with additional processing and annotations added for the purposes of the current work.
Each session contains a top-down video with synchronized audio, high-quality FLAC, and MIDI recordings.
We do not use the audio recordings in our experiments, but they are available for future work and for the benefit of the community.
We also provide tight crops around the piano keys for all videos.
Two different pianos are used throughout the practice sessions: a Yamaha GB1k Silent and an Essex EUP-116 equipped with a silent system (allowing for MIDI capturing).
For audio we use AKG P170 condenser microphones and a Focusrite Scarlett 2i2 audio interface.
The video is captured with GoPro cameras.

R3 includes two sub-splits which differ in content variety and size.
The R3s split contains practice sessions from one pianist practicing Rachmaninoff's Piano Concerto No.~3 Op.~30 (and hence the name of the dataset), while R3x contains sessions from both pianists practicing a variety of pieces from the Western Classical repertoire.
Both splits have been processed exactly the same way and differ only in musical content.

As part of our dataset preparation pipeline, we split all recordings into smaller chunks between $\sim50$ seconds and $\sim447$ seconds.
This results in 895 individual sets of synchronized video, FLAC, and MIDI recordings.
To create crops around the piano keyboard, we employ an automatic template matching approach with manual adjustments where necessary.
We create train and test sets using a 25/75 split, resulting in 181 samples in the combined test and 714 samples in the combined train set.
To ensure that each test set is comprised of completely different practice sessions from those present in the train sets, we ensure that no samples share the same date across splits.
Figure \ref{fig:dataset_distribution} shows the distribution of MIDI pitches throughout the datasets.

\subsection{PianoYT} 
Introduced by \cite{9053115}, this dataset consists of ca.~19.2 hours (shown in \tabref{tab:dataset_statistics}) of top-down piano performance videos downloaded from YouTube.
Dataset labels are generated automatically using the OaF model~\cite{DBLP:conf/ismir/HawthorneESRSRE18}.
We note that our version of the PianoYT dataset is missing 2 videos in the train set due to them no longer being available on YouTube.

The amount of notes and the total duration of both datasets are shown in \tabref{tab:dataset_statistics}.
Our R3 dataset has not only a longer total duration than PianoYT, but the total number of notes is more than double, indicating a higher note density than PianoYT. 
This could be attributed to either faster piano pieces being played or less silence throughout the R3 dataset.
PianoYT also has a significantly smaller test set than R3.

\section{Experimental Setup}\label{sec:experimental_setup}

\subsection{Video Pre-Processing}\label{sec:experimental_setup:video_preprocessing}

In order to make data loading easier and have a consistent temporal resolution throughout all datasets, all videos are first transcoded from their native frame-rate to a frame-rate of 30 using FFmpeg.\footnote{\url{https://ffmpeg.org/}}
We then crop out the piano using the crops provided with the datasets.
Lastly, we ensure that the image is in the correct orientation and is mirrored correctly.
Specifically, this requires rotating all PianoYT images 180 degrees to have the same orientation as R3.

\textbf{Video Augmentations}: 
As data augmentations we randomly apply Gaussian noise, random rotations, brightness and scale jittering.
We also utilize a grayscale transformation to remove potentially unimportant information from the image. 
Lastly, we normalize all frames using the mean and std. from ImageNet~\cite{5206848}.

\subsection{MIDI Pre-Processing}\label{sec:experimental_setup:midi_preprocessing}

When adding onset labels, we first assign each note onset to the nearest video-frame, giving it a label of $1$. 
Then we additionally assign a value of $0.5$ as an onset label for the next and previous video-frames. 
To calculate frame labels, we assign a label of $1$ when the note is pressed down, with the video-frame before and after the positive region being given a label of $0.5$. 
The additional positive values are added in order to increase the number of positive onset and frame values in the dataset and also to deal with possible differences in timing between datasets.

\subsection{Class Weights}\label{sec:class_weights}

We employ the binary cross-entropy (BCE) loss function, modeling our problem similarly to multi-label classification and conducting 176 binary classifications.
When training for joint onset and frame prediction, we obtain the final loss by summing the loss from both predictions.
Due to the sparsity of positive classes in the datasets, we also apply class weights directly to the output of the loss function, which gives us the ability to control the weight of the positive samples in the dataset. 
This results in the following loss function:

$$\mathcal{L}_{\text{BCE}} = -\frac{1}{N}\sum_{i=1}^{N}w_{i}\left[y_{i}\log(\hat{y}_{i}) + (1-y_{i})\log(1-\hat{y}_{i})\right] \label{bce_loss}.$$

\noindent where $y_i$ is the ground truth, $\hat{y}_i$ is the model output, and $w_{i}$ is the weight for the $i$-th sample, and $N$ is the number of samples.
By changing the positive class weights, we can balance the precision and recall of the model.

When choosing what class weights to use, we first calculate the ratio of positive to negative frames for each note.
We then rescale this to be within $[2, 4]$ for frames and $[3, 5]$ for onsets. 
Rescaling is something we found necessary, as we found very-high class weights led to unstable training.
Because onsets are sparser than frames, we use larger positive class weights for them.

\subsection{Training}\label{sec:experimental_setup:experimental_setup}

We utilize either two GPUs with a total batch size of 32 or 4 GPUs with a total batch size of 384 when training on PianoYT or R3 respectively.
When training on R3 we use the AdamW~\cite{DBLP:journals/corr/abs-1711-05101} optimizer with a weight decay of 0.05. 
When fine-tuning a model trained on R3 on PianoYT, we use SGD with a momentum of 0.9.
The learning rate for all runs follows a cosine decay function with a warmup of either 10\% or 25\% of all training steps depending on whether we are training on PianoYT or R3 respectively.
We utilize the same image augmentations as described in \secref{sec:experimental_setup:video_preprocessing} across all runs on R3 and PianoYT.

\subsection{Evaluation}\label{sec:experimental_setup:evaluation}

We follow the standard evaluation practices used in AMT \cite{DBLP:conf/ismir/HawthorneESRSRE18,9053115,hu2024towards,Ycart-2020} and 
we employ the precision, recall, and F-measure as our evaluation metrics.
These metrics can be computed either \emph{framewise}, where we consider each frame as a separate prediction, 
or \emph{notewise}, where each note is described by its onset, offset, and pitch. 
Additionally, notewise metrics can be computed \emph{onsetwise}, where we only consider the pitch and onset of each note, ignoring the offset,
which is useful for comparing with previous work that does not predict offsets.

For onsetwise and notewise metrics, we evaluate model performance using the transcription evaluation utilities provided by \verb|mir_eval|~\cite{DBLP:conf/ismir/RaffelMHSNLE14}. 
This implementation matches as many notes as possible by checking if the distance between predicted note onset/pitch and reference note onset/pitch is less than set tolerance values,
before computing the respective metrics using this matching.
We keep all settings at their defaults except the onset tolerance, which we increase from \SI{50}{\milli\second} to \SI{100}{\milli\second}.
This corresponds to an increase from an onset tolerance of $\sim\pm1$ to $\pm3$ frames, meaning that a predicted onset up to 3 frames away from the true onset would now be considered correct.
This window size in the \SI{100}{\milli\second} case is close to the limit of human detectability thresholds~\cite{1998RecIB}.
We show the effects different onset tolerances have on the model performance in \figref{fig:f_score_onset_tolerance}.
For framewise metrics, we follow~\cite{Ycart-2020,hu2024towards}, and we use a frame-rate of \SI{30}{\fps} to match the frame-rate of the input videos.

\section{Experimental Results}\label{sec:experiments}

\subsection{Augmentations and Class Weights}\label{sec:experiments:ablation}

\tabcolsep=0.07cm
\begin{table}[t]
  \centering
  \begin{tabular}{lccc}%
    \toprule
    \multicolumn{1}{c}{} & \multicolumn{3}{c}{F-score} \\
    Configuration & Framewise & Onsetwise & Notewise \\
    \midrule
    Color inputs       & 0.08 & 0.18 & 0.06 \\
    + Grayscale               & 0.12 & 0.25 & 0.07 \\
    + DataAugment       & 0.16 & 0.39 & 0.13 \\
    + Class Weights        & 0.31 & 0.63 & 0.20 \\
    \bottomrule
  \end{tabular}
  \caption{The effect of progressively adding augmentations and class weights on ViT performance.}
  \label{tab:ablation}
\end{table}

In order to explore the contribution of each design choice leading up to our final proposed ViT model training procedure, we perform an ablation study on the smaller R3s dataset.
To show the effect of preprocessing, data augmentation techniques, and class weights, we iteratively add them to the training recipe and observe the change in the models performance on the R3s validation set after training for 5 epochs.
More details on the exact hyperparameters used for the ablation study can be found in \apref{sec:ablation_hyper} in the supplementary materials.

Utilizing grayscale inputs already leads to an improvement in the model's performance, as is shown in \tabref{tab:ablation}.
Adding the collection of data augmentation techniques discussed in \secref{sec:experimental_setup:video_preprocessing}, indicated as DataAugment, as well as the class weights discussed in \secref{sec:class_weights}, continued to improve metrics greatly.
Overall, these results support our choice to use these techniques when training the ViT.

\subsection{Predictive Accuracy}\label{sec:experiments:predictive}

\begin{table*}[t]
\centering
\begin{tabular}{@{}lccc@{\hspace{20pt}}ccc@{\hspace{20pt}}c@{}}
\toprule
           & \multicolumn{3}{c}{R3s}             & \multicolumn{3}{c}{R3x}             & \multicolumn{1}{c}{PianoYT}    \\
Model      & Framewise & Onsetwise & Notewise & Framewise & Onsetwise & Notewise & Onsetwise \\ \midrule
\multicolumn{8}{l}{\textbf{Trained on R3}}                                                          \\
\multicolumn{8}{l}{Predicting Onsets and Frames}                                                    \\ \midrule
S2S~\cite{9053115}        & 0.31       & 0.84       & 0.18      & 0.46       & 0.83       & 0.46      & 0.06       \\
V2R~\cite{DBLP:conf/nips/SuLS20}       & 0.31       & 0.84       & 0.18      & 0.45       & 0.82       & 0.47      & 0.01       \\
ViT (ours) & 0.32       & 0.87       & 0.19      & 0.48       & 0.85       & 0.49      & 0.40       \\ \midrule
\multicolumn{8}{l}{Predicting Frames only}                                                          \\ \midrule
S2S~\cite{9053115}        & 0.51       & 0.82       & 0.27      & 0.63       & 0.81       & 0.55      & 0.24       \\
V2R~\cite{DBLP:conf/nips/SuLS20}        & 0.52       & 0.86       & 0.28      & 0.63       & 0.83       & 0.57      & 0.05       \\
ViT (ours) & 0.52       & 0.88       & 0.29      & 0.63       & 0.85       & 0.57      & 0.35       \\ \midrule
\multicolumn{8}{l}{\textbf{Trained on R3, finetuned on PianoYT}}                                    \\
\multicolumn{8}{l}{Predicting Onsets Only}                                                    \\ \midrule
S2S~\cite{9053115}        & -          & 0.21       & -         & -          & 0.30       & -         & 0.64       \\
V2R~\cite{DBLP:conf/nips/SuLS20}        & -          & 0.21       & -         & -          & 0.30       & -         & 0.64       \\
ViT (ours) & -          & 0.58       & -         & -          & 0.72       & -         & 0.68       \\ \midrule
\end{tabular}
\caption{Performance comparison (F-score) of models trained on R3 and fine-tuned on PianoYT}
\label{tab:predictive_accuracy}
\end{table*}

Here we show the performance of our ViT-based model on the R3 and PianoYT datasets and compare it to the previous ResNet-based models proposed by ~\cite{9053115,DBLP:conf/nips/SuLS20}.
We also investigate the differences between using the onsets-and-frames approach for note predictions versus the frames-only one to see which could be a better modeling strategy for this task.
This is done by training all models on joint onsets and frames, and then ignoring the onset predictions during post-processing.
All results are shown in \tabref{tab:predictive_accuracy}.
The model called S2S is our re-implementation of ResNet with aggregation and slope module presented by ~\cite{9053115}, and V2R is the official implementation provided by ~\cite{DBLP:conf/nips/SuLS20} for their Video2Roll Net.

\textbf{R3}: 
We train all models on joint onsets and frames on R3 for 10 epochs, utilizing all the methods tested in \secref{sec:experiments:ablation}.
When training our ViT model we additionally find it necessary to use stochastic depth~\cite{DBLP:conf/eccv/HuangSLSW16} to combat overfitting.
When training the CNNs we did not encounter overfitting to the same degree as with the ViT, and instead chose to use use early stopping.
Additional details on the parameters for these runs can be found in \apref{sec:pianoyt_hyper} in the supplementary material.
We evaluate metrics on the two subsplits of R3, R3s and R3x, as well as on PianoYT, which is not included in the training data.

In all metrics, our ViT model outperforms the other two.
Interestingly, when looking at the performance on PianoYT, the ViT is the only model that somewhat generalizes to the unseen dataset.
We also find interesting that, although there are some architectural differences between the two CNNs, they both perform very similarly across a wide range of metrics. 
This suggests that the most important factor for the performance of these models is the shared ResNet backbone.

When comparing metrics between onsets-and-frames and frames-only prediction approaches, the first approach seems to fall behind.
Although onsetwise metrics are very similar for the two approaches, framewise and notewise metrics are higher when predicting only frames.
Framewise scores seem to benefit the most when utilizing frame-only predictions.

\textbf{PianoYT}:
When training on PianoYT, we utilize the models that were already pre-trained on R3 and then fine-tune them on PianoYT.
For this reason, we find it sufficient to train for only 5 epochs on PianoYT.
Detailed hyper-parameters for these runs can be found in \apref{sec:pianoyt_hyper} in the supplementary material.
Since the PianoYT ground truth offsets are of low quality, we only train on onsets.
Our ViT model shows better performance again, while both ResNet-based models achieve the same onsetwise scores.
Perhaps the largest difference between the ViT and ResNet models is the ability of the ViT to better maintain performance on the R3 dataset it was pre-trained on.

\section{Discussion}\label{sec:discussion}

Our results across all datasets show strong scores in favor of the ViT based model.
Our model is able not only to outperform previous ResNet based approaches on the data it was trained on, but also shows very promising generalization capabilities. 
We believe that these capabilities are the result of a combination of high quality pre-trained weights and a larger model that is able to learn and retain more information.
The large size of R3 is also likely more beneficial to a large model that has the capacity to learn from the extra data.

We also discovered that a frames-only approach works better than joint onsets and frames.
This could be due to the sparsity of onset events as compared to frame-level events, which could lead to poorer quality onset predictions.
These results show a stark difference from audio based models, which have been shown to benefit from the joint modeling task.
We did find the onset predictions useful when fine-tuning on PianoYT however, as they allowed us to adapt the joint onset and frame model to just onsets without having to make any changes.

\begin{figure}[t]
\centering
\includegraphics[width=\linewidth]{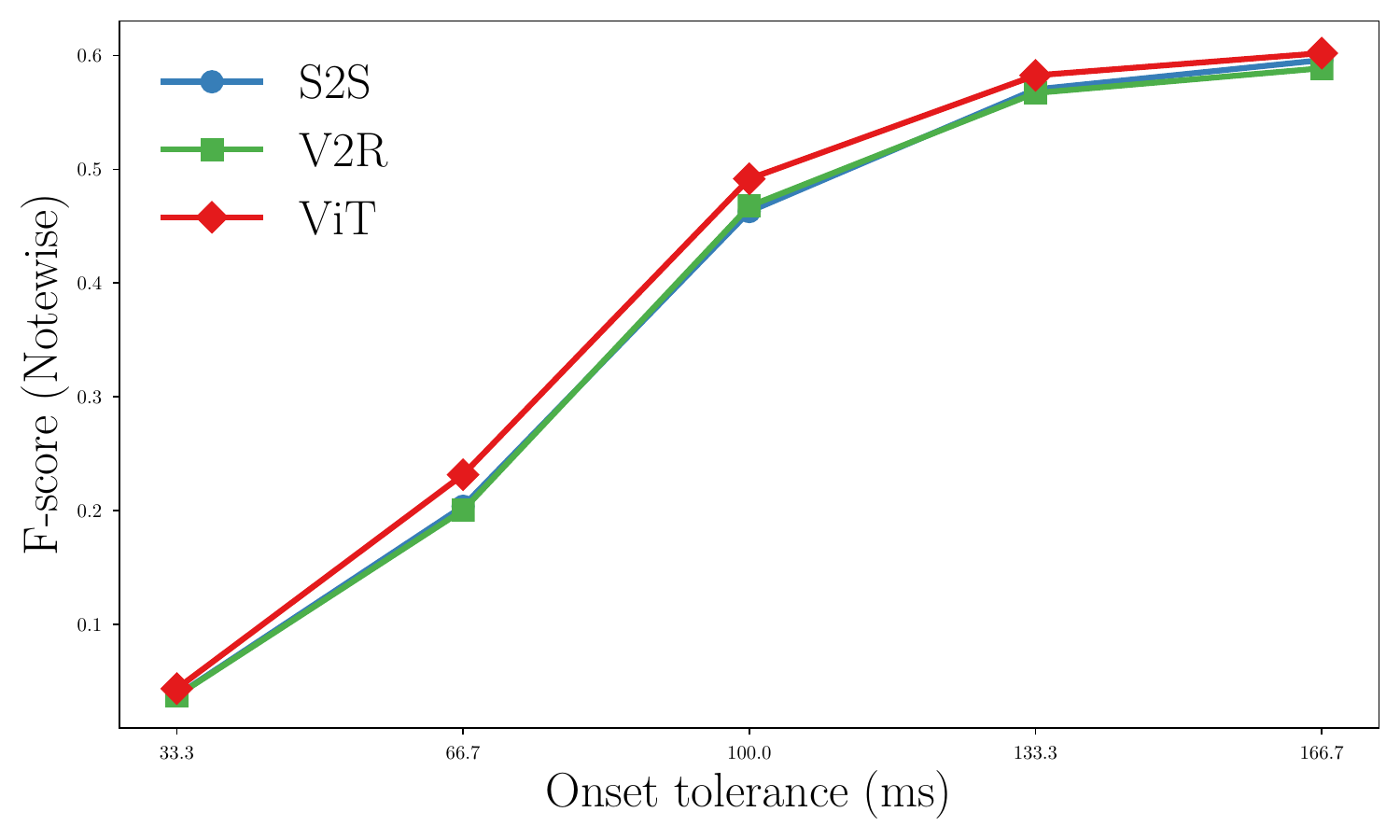}
\caption{Notewise F-score on R3x for different onset tolerances for ViT, S2S and V2R models.}
\label{fig:f_score_onset_tolerance}
\end{figure}

\paragraph{Onset Tolerances.}
To investigate the temporal accuracy of the models on the R3x dataset, we plot the notewise F-score in \figref{fig:f_score_onset_tolerance}.
Although all models struggle to predict notes within a one frame tolerance, they get much better up to a 3 frame tolerance, at which point they start to plateau.
These results show how challenging it is to make temporally accurate predictions at a frame-rate of \SI{30}{\fps}, and highlight an area where VPT algorithms can still be greatly improved.

\paragraph{Supplementary Materials.}
We provide a Google Colab showcasing the inference procedure of our model and a YouTube playlist with example predictions at this web page.\footnote{\url{https://chromeilion.github.io/onf_vpt/}}
Dataset, model weights, and training/evaluation code are also public and made available.

\section{Conclusions}\label{sec:conclusions}
This paper introduces a transformer architecture for visual piano transcription.
Our experimental results show that the proposed architecture can compete with and even outperform previous CNN-based models.
To further aid model training, we introduce a new dataset, R3, containing synchronized video, audio and MIDI data.
Additionally, we investigate the effect of positive class weights on model performance, highlighting their importance and impact on the final model.
We also compare two approaches for note onset and offset prediction and discover that the frames-only approach could provide better results than the joint onsets-and-frames one
for VPT in contrast to audio-based AMT, highlighting the differences between the two tasks.

Future research could focus on improving temporal accuracy by training on a higher frame rate or by making sub-frame predictions. 
Another possible direction is addressing the strong influence of the model's pre-trained weights on our method, potentially by conducting pre-training specifically for the VPT task. 
We would also like to investigate explicitly learning  offsets, similarly to the approach used for onsets, to obtain better performance in notewise and framewise metrics.
Furthermore, a full transformer Encoder--Decoder could be used to predict MIDI-like tokens directly from the video clip, using a larger temporal size, such as 32 frames, to process approximately one second of video at a time. 
This approach would function similarly to \cite{DBLP:conf/ismir/HawthorneSSME21} and eliminate the need for a hand-crafted post-processing pipeline. 
Finally, more work in the direction of utilizing both audio and video for AMT could lead to overcoming the limitations of each of the modalities
and achieving even better results.

\clearpage
\section*{Acknowledgments}

This work has been supported by the Austrian Science Fund (FWF), grant agreement PAT 8820923 (``\emph{Rach3: A Computational Approach to Study Piano Rehearsals}'').
A portion of the computational resources this work utilizes were provided by the Italian AREA Science Park supercomputing platform ORFEO in Trieste.

\bibliographystyle{named}
\bibliography{ijcai25}

\newpage
\begin{appendices}

\maketitle
\appendix
\section{Ablation Hyperparameters}
\label{sec:ablation_hyper}

All experiments in the ablation study were performed with our ViT model trained on the smaller R3s subset and used the hyperparameters listed in Table~\ref{tab:ablation_training} for training. Configurations that use DataAugment utilize all augmentations listed in Table~\ref{tab:dataaug}.

\tabcolsep=0.07cm
\begin{table}[h!]
  \centering
  \begin{tabular}{lcc}%
    \toprule
                      &  R3s \\
    \midrule
    Optimizer         & AdamW  \\ 
    Weight decay      & 0.05  \\
    Learning rate     & 1$\times10^{-3}$  \\
    Warmup percentage & 0.25   \\
    Epochs            & 5  \\
    Batch Size  & 96   \\
    
    \bottomrule
  \end{tabular}
  \caption{Training hyperparameters for all ablation experiments.}
  \label{tab:ablation_training}
\end{table}

\section{R3 and PianoYT Training Hyperparameters}
\label{sec:pianoyt_hyper}

The hyperparameters used when training our ViT model are detailed in Table \ref{tab:ViT_R3}. 
Hyperparameters for the S2S and Video2roll models are detailed in Table \ref{tab:S2S_R3}.
All these runs utilize the image augmentations detailed in Table \ref{tab:dataaug}.

\tabcolsep=0.07cm
\begin{table}[ht!]
  \centering
  \begin{tabular}{lccc}%
    \toprule
                      & R3    & PianoYT  \\
    \midrule
    Optimizer         & AdamW & SGD \\ 
    Momentum          &  -    & 0.9 \\ 
    Weight decay      & 0.05  & 0.0 \\
    Learning rate     & 1$\times10^{-3}$ & 0.25 \\
    Warmup percentage & 0.25  & 0.10  \\
    Epochs            & 10    & 5  \\
    Total Batch Size  & 384   & 32 \\
    Stochastic droplayer rate & 0.3 & 0.0 \\
    Pytorch AMP     & yes & yes \\
    \bottomrule
  \end{tabular}
  \caption{Training hyperparameters for the ViT model from the main paper.}
  \label{tab:ViT_R3}
\end{table}

\tabcolsep=0.07cm
\begin{table}[ht!]
  \centering
  \begin{tabular}{lccc}%
    \toprule
                      & R3    & PianoYT  \\
    \midrule
    Optimizer         & AdamW & SGD \\ 
    Momentum          &  -    & 0.9 \\ 
    Weight decay      & 0.05  & 0.0 \\
    Learning rate     & 1$\times10^{-3}$ & 0.25 \\
    Warmup percentage & 0.25  & 0.10  \\
    Epochs            & 10    & 5  \\
    Total Batch Size  & 384   & 32 \\
    Stochastic droplayer rate & 0.0 & 0.0 \\
    Pytorch AMP     & no & no \\
    \bottomrule
  \end{tabular}
  \caption{Training hyperparameters for both the S2S and Video2roll model.}
  \label{tab:S2S_R3}
\end{table}

\tabcolsep=0.07cm
\begin{table}[h!]
  \centering
  \begin{tabular}{lccc}%
    \toprule
                      &  Paremeters & Probability \\
    \midrule
    ScaleJitter\tablefootnote{\url{https://docs.pytorch.org/vision/main/generated/torchvision.transforms.v2.ScaleJitter.html}}       & min: 0.96, max: 1.001 & 1 \\ 
    ColorJitter\tablefootnote{\url{https://docs.pytorch.org/vision/main/generated/torchvision.transforms.v2.ColorJitter.html}}       &  brightness: 0.1     & 0.4 \\ 
    RandRotation\tablefootnote{\url{https://docs.pytorch.org/vision/main/generated/torchvision.transforms.v2.RandomRotation.html}}      & degrees: 0.2  & 0.4 \\
    Grayscale         & -  & 1 \\
    \bottomrule
  \end{tabular}
  \caption{Data augmentation details, these settings were used across all training runs on R3 and PianoYT.}
  \label{tab:dataaug}
\end{table}

\section{Additional Implementation Details}
\label{sec:implementation_details}

Here we provide additional information on some of the techniques we used when training the ViT model.

\textbf{Pytorch AMP:}

In order to speed up some training instances of our ViT models on the A40 GPUs, we utilized half-precision (float16) training handled by Pytorch's Automatic Mixed Precision\footnote{\url{https://docs.pytorch.org/docs/stable/amp.html}} (AMP) package. These instances are respectively marked in Table~\ref{tab:ViT_R3}.

\textbf{Stochastic Droplayer:}
Stochastic droplayer is a form of dropout where whole layers in the model are zeroed out with the following probability:

\begin{equation}
    p_{l}=\frac{l}{L}p_{L}
\end{equation}

Where $p_{l}$ is the probability of the layer at depth $l$ being zeroed out, $p_{L}$ is the probability of the last layer being zeroed out, and $L$ is the number of layers in the model.
Therefore, the probability $p_{l}$ grows linearly as depth increases until it reaches the maximum $p_{L}$.

\end{appendices}

\end{document}